\documentclass[letterpaper, 10 pt, conference]{ieeeconf}  

\IEEEoverridecommandlockouts                              

\usepackage{amsmath} 
\usepackage{float}
\usepackage{amsfonts} 

\usepackage{tikz}
\usetikzlibrary{shapes, arrows, positioning}
\usepackage{makecell} 

\title{\LARGE \bf
Assistive Decision-Making for Right of Way Navigation \\ at Uncontrolled Intersections}

\author{
    Navya Tiwari$^{*}$, Joseph Vazhaeparampil$^{*}$, Victoria Preston%
    \thanks{$^{*}$These authors contributed equally to this work.}%
    \thanks{All authors are with Olin College of Engineering, Needham, MA 02492, USA
    {\tt\small \{ntiwari, jvazhaeparampil, vpreston\}@olin.edu}}%
}

\begin{document}

\maketitle
\thispagestyle{empty}
\pagestyle{empty}


\section{INTRODUCTION}
Intersections account for nearly 40\% of U.S. crashes \cite{NHTSA2023}, with many occurring at uncontrolled or partially controlled locations \cite{Noyce2002}. Ambiguous right of way, compounded by occlusions, non-compliant drivers, and limited sensing leave drivers uncertain how to act \cite{NSC2022}. Addressing these challenges requires assistive technology that reduces driver uncertainty and improves awareness. We propose an Advanced Driver Assistance System (ADAS) that fuses sensor data, interprets intersection context, and applies uncertainty-aware frameworks to recommend safe actions at uncontrolled intersections. We pose three research questions (RQs):

\begin{enumerate}
\item[RQ1] How can the accuracy of ego- and external-vehicle state estimation be improved while constraining uncertainty through frustum-based fusion of camera and lidar data, given real-time, computationally limited resources?
\item[RQ2] To what extent can a driver-assist system enable safe navigation of uncontrolled intersections by efficiently interpreting intersection context (lane markings, stop signs, traffic flow patterns, and pedestrian presence) under partial observability?
\item[RQ3] How effectively can different decision-making frameworks handle uncertainty at uncontrolled intersections, and what trade-offs emerge between safety, efficiency, and computational feasibility in real-time deployment?
\end{enumerate}

Here, we present an initial analysis of RQ3 in synthetic uncontrolled intersections. We show that probabilistic planners, particularly POMCP (Partially Observable Monte Carlo Planning) \cite{Silver2010} and DESPOT (Determinized Sparse Partially Observable Tree) \cite{Somani2013}, outperform deterministic approaches in predicting the intent of other drivers and selecting collision-free actions, while maintaining safety under complex, partially observable scenarios. Continued work will integrate Sensor Fusion (RQ1) and Environment Perception (RQ2) modules for end-to-end, real-time navigation under realistic traffic and environmental conditions.

\section{BACKGROUND AND RELATED WORK}
Prior works on rule-based systems \cite{LeFevre2010,Wray2021} and reinforcement learning \cite{Kyrki2021,Sunberg2017} have advanced intersection management through structured or learned policies but struggle in scenarios outside their training data \cite{Bai2014, Arkin2006}. Uncertainty-aware frameworks, such as decision making algorithms with Partially Observable Markov Decision-Processes (POMDPs), make steps in this direction \cite{Thornton2022} but typically assume full autonomy authority for the ego vehicle, which most vehicles currently lack \cite{NHTSA2023,NSC2022}. As such, a clear gap persists in systems that guide drivers without replacing them. To realize safer roads in the near future requires an ADAS to retrofit existing vehicles, leverage human-robot interaction practices \cite{Goodrich2007}, and operate on accessible, low-cost sensors \cite{BenElallid2023,Kolter2012}. Such ADAS research remains limited, and existing ADAS tends to prioritize collision avoidance and lane keeping \cite{Bai2014} over right of way reasoning at intersections. Our work aims to address this gap by developing a prototype assistive navigation framework for right-of-way reasoning at uncontrolled intersections.

\section{Problem Formulation}
We formalize the Right of Way navigation task as a POMDP defined by the tuple
$M = \langle S, A, T, R, \Omega, O, \gamma, b_0 \rangle$:

\begin{itemize}
    \item State space $S = \{ r_n, p_n, s_\text{ego} \}_{n=1}^N$, where $r_n \in \mathbb{Z}^+$ represents the arrival order of the $n$-th other vehicle, $p_n \in \{0,1\}$ indicates the presence of a pedestrian in the path of the $n$-th vehicle, $s_\text{ego} \in \mathbb{R}^d$ encodes the state of the ego vehicle (position, velocity, heading, etc.), and $N$ is the total number of other vehicles at the intersection.
    \item Action space $A = \{\text{STOP, YIELD, GO}\}$ encodes recommended maneuvers for the driver of the ego vehicle.
    \item Transition function $T(s' \mid s, a) = P(s_{t+1} = s' \mid s_t = s, a_t = a)$ captures stochastic evolution of vehicle ordering and pedestrian presence, including occasional non-compliant driver behavior.
    \item Reward function $R(s, a) = r_\text{safety} + r_\text{efficiency}$ where $r_\text{safety} < 0$ penalizes unsafe maneuvers and collisions and $r_\text{efficiency} > 0$ rewards safe and efficient passage.
    \item Observations $\Omega = \{ o_n \}_{n=1}^N, \quad o_n \in \mathbb{R}^k$ are noisy and partial measurements drawn from observation model $O(o \mid s, a) = P(o \mid s, a)$.
    \item Discount factor $\gamma \in [0,1]$ .
    \item Belief $b(s) = P(s \mid h_t)$, where $h_t = \{a_0, o_1, \dots, a_{t-1}, o_t\}$ is the history of actions and observations. The belief state represents a probability distribution over all possible states and is updated using a Bayesian filter.
\end{itemize}

Directly solving the Right of Way POMDP is generally intractable \cite{Bai2014_2}, so an approximate solver must be employed. We present our own compositional framework for solving the Right of Way POMDP, considering the challenges of cooperative driving, in the following section.

\section{Methodology} \label{m}
In right of way navigation, an ego-vehicle is presented with the choice to come to a complete stop, yield, or proceed on their intended trajectory, based on the rules of the road and the actions of other drivers. We investigated the efficacy of four planners (approximate solvers for the right of way POMDP) --- an FSM, QMDP, POMCP, and DESPOT --- using a custom Python-based testing workbench with synthetic uncontrolled four-way intersections that simulate randomized vehicle behaviors, pedestrian presence, and environmental uncertainties. A common software wrapper enforced uniform logging, applied perception latency, and ensured comparable evaluation across planners.

The FSM was implemented from scratch as a deterministic baseline. It encodes traffic-law heuristics, prioritizing pedestrians, earlier arrivals, right-hand vehicles on ties, and left-turn conflicts before permitting a “Go” action. Intent prediction is approximated using a heuristic over vehicle kinematics and intersection context, estimating whether each agent is likely to yield, proceed, or turn. The three probabilistic planners were implemented using specialized solver packages in combination with numerical libraries. QMDP was developed with NumPy \cite{numpy} to perform value iteration over a compact POMDP state space, converging to a stationary policy applied to belief distributions. POMCP was adapted from algorithms available in the JuliaPOMDP framework \cite{juliapomdp}, with a Python implementation that combines NumPy \cite{numpy} for efficient particle sampling and pomdp-py \cite{pomdp-py} utilities for belief management and rollout simulation. DESPOT likewise drew on JuliaPOMDP’s DESPOT implementation \cite{juliapomdp_despot} as a reference, with NumPy \cite{numpy} used for particle evaluation and aggregation.
 We configured POMCP with 150 simulations and a planning depth of six, while DESPOT was run with 32 particles and depth four, reflecting typical trade-offs between accuracy, efficiency, and real-time feasibility.

Scenarios were generated synthetically to reflect realistic complexity. Each scenario specifies an ego vehicle with a weighted intent distribution (55\% straight, 30\% left, 15\% right), up to three other vehicles with randomized arrival times, and pedestrians with a 12–25\% probability of appearance. Additional stochasticity is introduced through Gaussian sensor noise ($\sigma$ = 1 m, ±1 timestep arrival error), occlusions (15\%), perception dropouts (2\%), and slippery road conditions (12\%). Roughly one-third of scenarios were adversarial: increased frequency of simultaneous arrivals, stop-running vehicles, and pedestrian intrusions. We evaluated all planners across the same 60 scenarios with a 12-step horizon. Metrics included action accuracy, intent prediction accuracy, collision-free rate, traffic flow efficiency, time-to-completion, real-time feasibility (deadline 0.05 s), and near-miss recovery. real-time

\section{Preliminary Results}

The FSM baseline underperformed in complex traffic, with less than 60\% action and intent accuracy across the 60 scenarios, with very low performance when encountering simultaneous arrivals and partial observability. 

In contrast, QMDP achieved 78.3\% action accuracy and 86.2\% intent prediction accuracy, striking a balance between computational efficiency and performance. DESPOT maintained a 93.3\% collision-free rate while completing tasks efficiently, often outperforming POMCP in runtime. POMCP demonstrated the highest safety, with a collision-free rate of approximately 97.5\%, but incurred higher computational costs, averaging 0.03 – 0.05 s per decision compared to 0.001 – 0.002 s for FSM and QMDP. Figure 1 and Table 1 summarize the results of the initial investigation.

\begin{figure} [H]
        \centering
        \includegraphics[width=1\linewidth]{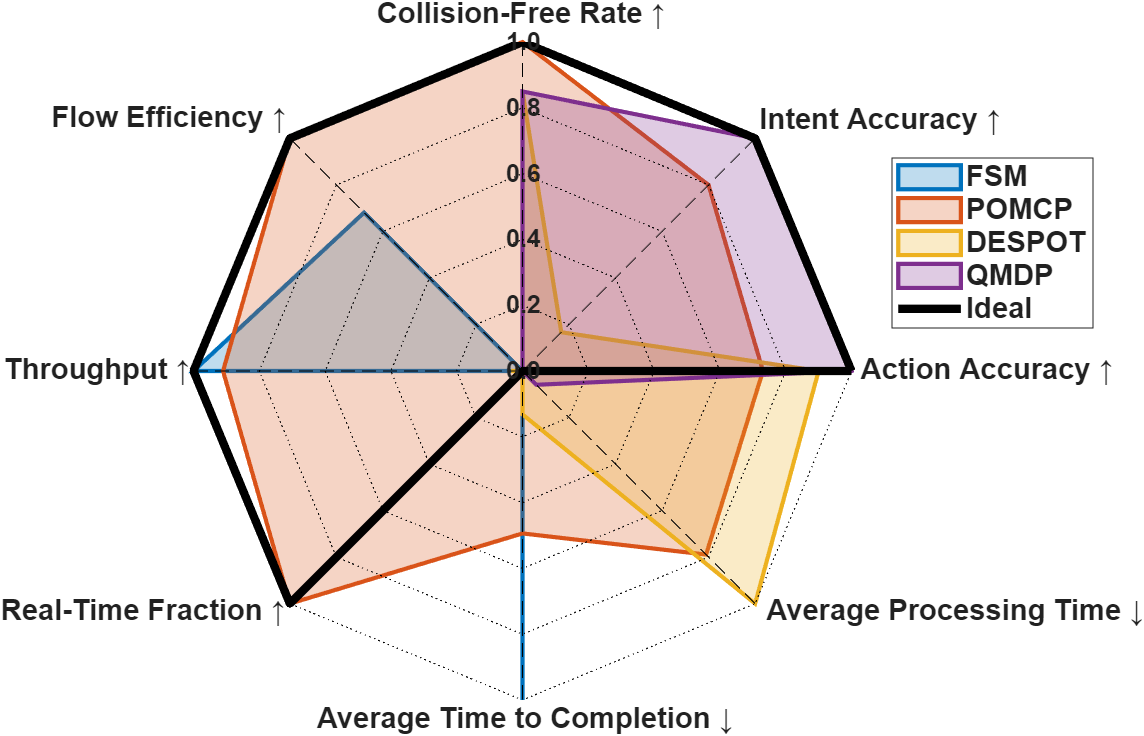}
        \caption{Radar chart showing trade-offs in planner performance across eight normalized metrics. Metrics include action accuracy (percentage of ego-vehicle maneuvers correctly executed as intended), intent accuracy (percentage of correctly predicted future actions of other vehicles in the scene), collision-free rate (fraction of simulation runs completed without any collisions), flow efficiency (ratio of vehicles passing the intersection ideally), average time to completion (mean time taken for all vehicles to completely clear the intersection), and average processing time (mean computation time required per decision step by the planner). Each metric is min–max normalized, where 0 corresponds to the lowest and 1 to the highest observed value.}
        \label{fig:enter-label}
\end{figure} 
\vspace{-2em}
\begin{table}[h]
    \caption{Summary of planner performance across 60 simulated intersection scenarios. Metrics highlight differences in accuracy, safety, efficiency, and computational cost.}
    \centering
    \begin{tabular}{c|c|c|c|c}
        Metric & FSM & QMDP & POMCP & DESPOT \\
        \hline
        $\uparrow$ Action Accuracy (\%) & 61.7 & \textbf{78.3} & 73.8 & 76.7 \\
        $\uparrow$ Intent Accuracy (\%) & 84.3 & \textbf{86.2} & 85.8 & 84.6 \\
        $\uparrow$ Collision-Free Rate (\%) & 61.7 & 93.3 & \textbf{98.9} & 93.3 \\
        $\uparrow$ Flow Efficiency (\%) & 40.0 & 35.0 & \textbf{42.3} & 35.0 \\
        $\downarrow$ Avg. Time to Comp. (s) & 1.04 & \textbf{0.50} & 0.77 & 0.57 \\
        $\downarrow$ Avg. Process. Time (s) & \textbf{0.0203} & 0.0204 & 0.0218 & 0.0222 \\
        $\uparrow$ Throughput & \textbf{0.390} & 0.323 & 0.384 & 0.325 \\
        $\uparrow$ Real-Time Fraction & 1.000 & 1.000 & \textbf{1.024} & 1.000 \\
    \end{tabular}
    \label{tab:planner_performance}
\end{table}

\vspace{-1.8em}

\section{Conclusion and future work}
Our study demonstrates that uncertainty-aware planners outperform deterministic approaches in right of way decision management at uncontrolled intersections, while also highlighting the significant impact of the form of uncertainty representation on overall planner performance. Future work will further characterize the role of uncertainty in planning, and will expand the implementation to Sensor Fusion (RQ1) and Environment Perception (RQ2) modules, with attention towards more complex, real-world scenarios and real-time deployment.





\bibliographystyle{unsrt}

\clearpage
\section{Supplemental Information}
\subsection{Expanded Background Discussion}
Uncontrolled intersections present various hazards due to ambiguous right-of-way rules, varying traffic patterns, and human driver unpredictability, contributing to a substantial portion of crashes in the U.S. \cite{NHTSA2023,Noyce2002}. Unlike signalized intersections, these locations rely on informal heuristics, such as yielding to the right or following arrival order, which may be interpreted inconsistently by different drivers \cite{NSC2022}. This ambiguity not only increases the risk of accidents, but also reduces traffic efficiency by causing congestion and delays. While autonomous vehicles are currently being developed to navigate such scenarios, most vehicles on the road today remain human-operated, highlighting the need for assistive systems that enhance situational awareness and guide driver decisions without requiring full autonomy \cite{Goodrich2007,Arkin2006}.

Decision-making under uncertainty is central to navigating intersections safely. Partially Observable Markov Decision Processes (POMDPs) provide a formal framework for modeling stochastic behaviors of vehicles, pedestrians, and sensor noise \cite{Bai2014,Somani2013}. Belief states allow planners to reason probabilistically about possible configurations of the environment, enabling informed decisions even with incomplete observations \cite{Ross2008}. Deterministic rule-based methods, such as finite state machines, offer simplicity and interpretability, but cannot fully capture rare or complex interactions \cite{LeFevre2010,Kyrki2021}. By constrast, probabilistic planners explicitly account for uncertainty, providing a more robust balance between safety, efficiency, and real-time computational feasibility \cite{Bai2014,Thornton2022}.

Existing research explores heuristic, reinforcement learning, and POMDP-based approaches for intersection navigation. Rule-based systems are interpretable but limited to pre-defined scenarios \cite{LeFevre2010,Wray2021}, while reinforcement learning methods can adapt to new situations but require extensive data and careful training \cite{Sunberg2017,Bouton2019}. POMDP-based methods excel in modeling uncertainty and interactions, particularly in fully autonomous contexts \cite{Bai2014,Hubmann2018}, yet few studies focus on assistive frameworks for human drivers. This gap motivates the development of driver-assist systems that integrate probabilistic planning with human-compatible outputs and low-cost sensors \cite{BenElallid2023,Kolter2012}.

Practical assistive systems must operate under constraints such as limited sensing, noise, occlusions, and computational budgets. Camera and lidar sensors provide complementary information, but are subject to environmental limitations and perception errors \cite{Cho2014,Kolter2012}. Moreover, human drivers are the final decision-makers, necessitating interfaces that communicate guidance clearly to avoid confusion or over-reliance \cite{Goodrich2007}. Integrating human-robot interaction principles into assistive design ensures that probabilistic recommendations are interpretable and actionable in real-world driving conditions \cite{Arkin2006,Thornton2022}.

\subsection{Detailed Research Questions}
\begin{enumerate}
\item[RQ1] \textbf{How can the accuracy of ego- and external-vehicle state estimation be improved while constraining uncertainty through frustum-based fusion of camera and lidar data, given real-time, computationally limited resources?} Accurate estimation of the ego vehicle and surrounding vehicles is critical for safe decision-making at intersections. Sensor noise, occlusions, and limited field-of-view introduce uncertainty in measurements. Frustum-based fusion of camera and lidar data can focus computation on relevant regions and reduce the impact of outliers. Efficient probabilistic filtering or particle-based approaches can balance estimation accuracy with real-time computational constraints \cite{Cho2014, Kolter2012}.

\item[RQ2] \textbf{To what extent can a driver-assist system enable safe navigation of uncontrolled intersections by efficiently interpreting intersection context (lane markings, stop signs, traffic flow patterns, and pedestrian presence) under partial observability?} Driver-assist systems must interpret intersection context under partial observability. Lane markings, stop signs, traffic flow, and pedestrian presence provide essential cues for right-of-way decisions. Sensor fusion combined with probabilistic reasoning can generate robust estimates of the environment \cite{Bai2014, Arkin2006}. Predictive models of other vehicles and pedestrians help the system guide drivers safely, even in unusual or complex traffic situations.

\item[RQ3] \textbf{How effectively can different decision-making frameworks handle uncertainty at uncontrolled intersections, and what trade-offs emerge between safety, efficiency, and computational feasibility in real-time deployment?} Different frameworks for decision-making have trade-offs in safety, efficiency, and computation. Rule-based approaches are fast but may fail in complex scenarios. Probabilistic planners such as QMDP, POMCP, and DESPOT account for uncertainty and partial observations, improving action accuracy and collision avoidance \cite{Bai2014, Somani2013, Thornton2022}. Evaluating these planners helps identify methods that provide reliable guidance while remaining feasible for real-time driver assistance.

\end{enumerate}
\subsection{Additional Figures}
\textbf{Definitions of Evaluation Metrics Used for Planner Performance:}
\begin{itemize}
    \item \textbf{Action Accuracy:} Percentage of times the planner selects the correct driving maneuver (stop, go, yield).
    \item \textbf{Intent Accuracy:} Percentage of times the planner correctly infers the intentions of other vehicles.
    \item \textbf{Collision-Free Rate:} Fraction of simulated runs completed without collisions.
    \item \textbf{Flow Efficiency:} Ratio of vehicles that traverse the intersection without experiencing extra waiting or slowing beyond what is required for safe and rule-compliant movement, compared to the optimal traffic flow.
    \item \textbf{Average Time to Completion:} Average total time required for all vehicles to clear the intersection in a scenario.
    \item \textbf{Average Processing Time:} Mean computational runtime per decision step for the planner.
    \item \textbf{Throughput:} Number of vehicles successfully processed through the intersection per unit time.
    \item \textbf{Real-Time Fraction:} Fraction of decision steps where the planner completed computation within the real-time deadline.
\end{itemize}

To evaluate planner performance, we defined a set of quantitative metrics: action accuracy, intent accuracy, collision-free rate, flow efficiency, average time to completion, and average processing time. These metrics collectively capture safety, efficiency, and computational feasibility. Using these definitions, we generated comparative graphs across planners, which highlight trade-offs between safety, efficiency, and runtime. The visualizations allow us to directly contrast planner behavior under identical intersection scenarios and identify which approaches achieve more balanced performance.

\begin{figure}[H]
    \centering
    \includegraphics[width=0.48\textwidth]{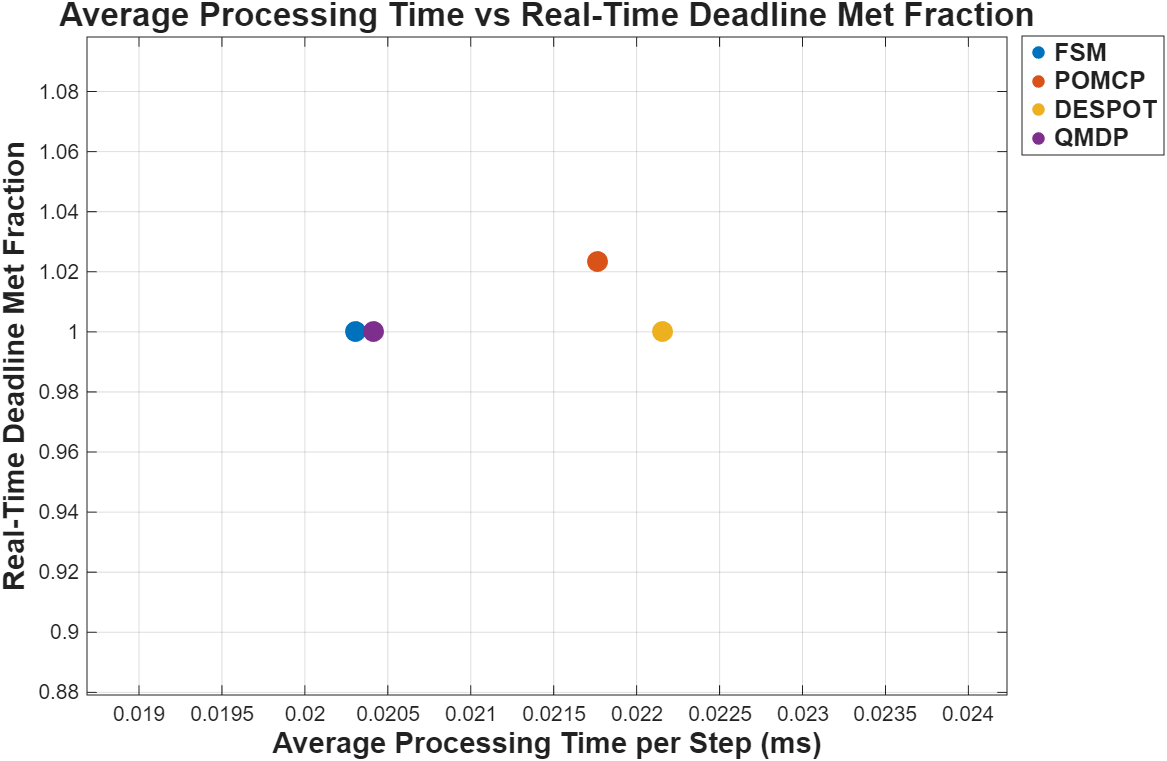} 
    \caption{Relationship between the planner’s average computation time per decision step and the fraction of steps completed within the real-time deadline. Each point represents the mean over 60 simulation runs, with crosses indicating variance in performance. All planners meet real-time requirements, with processing times comfortably below the deadline. Dashed lines indicate the real-time deadline for reference.}
    \label{fig:proc_time}
\end{figure}
Figure \ref{fig:proc_time} examines the relationship between Average Processing Time per Step and the Real-Time Deadline Met Fraction. All planners meet or exceed the real-time deadline requirements, with average processing times well below the threshold, ensuring timely decision-making suitable for real-world deployment.

\begin{figure}[H]
\centering
\begin{tikzpicture}[auto, node distance=4cm, thick, scale=0.8, transform shape]
\node (fusion) [draw, rectangle, rounded corners, minimum height=0.8cm, minimum width=2.5cm] {\makecell{Sensor\\Fusion}};
\node (perception) [draw, rectangle, rounded corners, minimum height=0.8cm, minimum width=2.5cm, right of=fusion] {\makecell{Environment\\Perception}};
\node (logic) [draw, rectangle, rounded corners, minimum height=0.8cm, minimum width=2.5cm, right of=perception] {\makecell{Right-of-Way\\Logic}};
\draw[->, thick] (fusion) -- (perception);
\draw[->, thick] (perception) -- (logic);

\end{tikzpicture}
\caption{Flowchart of the driver-assist system showing the sequential processing from Data input through Sensor Fusion, Environment Perception, and Right-of-Way Logic, leading to the system Output.}
\label{fig:spaced_flowchart}
\end{figure}
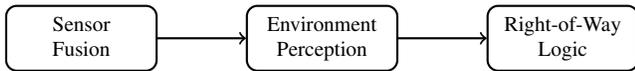
Figure \ref{fig:spaced_flowchart} illustrates the overall architecture of the driver-assist system. Raw data from sensors enters the system, which is first processed by the Sensor Fusion module to combine camera and lidar information. The fused data is then analyzed by the Environment Perception module to interpret intersection context, including lane markings, traffic participants, and pedestrian presence. Finally, the Right-of-Way Logic module uses this information to determine safe and efficient driving actions, producing the system Output for driver guidance.

While Section~\ref{m} describes how scenarios are generated, it is also useful to consider concrete examples. 
Table~\ref{tab:adv_examples} illustrates two representative cases directly produced by our simulator. 
The non-adversarial example depicts a simple interaction where the ego vehicle proceeds safely due to clear right of way. 
\begin{table}[h]
    \caption{Examples of non-adversarial and adversarial scenarios generated by the code.}
    \centering
    \renewcommand{\arraystretch}{1.1} 
    \setlength{\tabcolsep}{4pt} 
    \begin{tabular}{p{0.28\linewidth} p{0.65\linewidth}}
        \hline
        \textbf{Scenario Type} & \textbf{Description (actual generated examples)} \\
        \hline
        Non-Adversarial & Ego vehicle arrives from the north intending to go straight. One other vehicle approaches from the east, arriving 2 seconds later. No pedestrians are present, dry road conditions. Ground-truth safe action: \texttt{GO}. \\
        \hline
        Adversarial & Ego vehicle arrives from the west intending to turn left. Two other vehicles arrive simultaneously (from north and south), with one running the stop sign. A pedestrian suddenly emerges from occlusion mid-scenario, and the road is slippery. Ground-truth safe action: \texttt{YIELD}. \\
        \hline
    \end{tabular}
    \label{tab:adv_examples}
\end{table}

The adversarial example highlights a complex edge case, where multiple simultaneous arrivals, a non-compliant driver, a sudden pedestrian, and adverse road conditions all combine to challenge planner robustness. 
These contrasting cases demonstrate how the scenario generator stresses planners with both routine and difficult situations, ensuring that evaluation covers the full range of possible intersection dynamics.

\begin{figure}[h]
    \centering
    \includegraphics[width=0.85\linewidth]{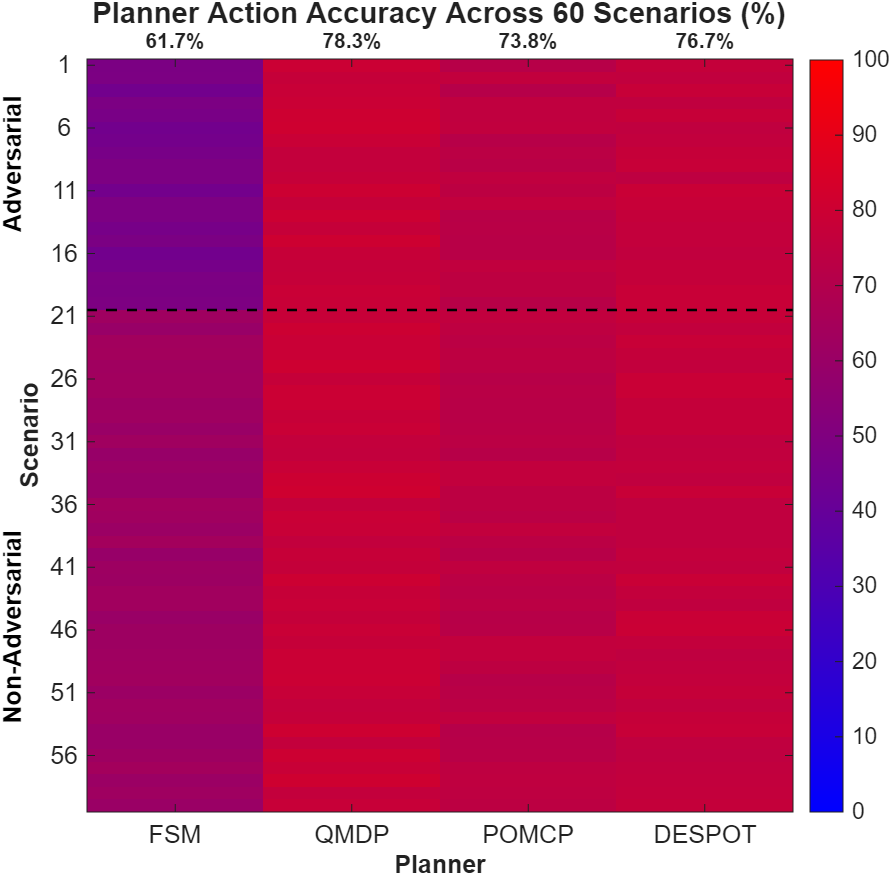}
    \caption{Planner action accuracy across 60 generated scenarios. Rows correspond to individual scenarios, with the first twenty marked as adversarial (above the dashed line). Columns correspond to planners. Each scenario consists of multiple timesteps, and the accuracy percentages for a scenario are computed based on the fraction of timesteps in which the planner selected the correct action. FSM shows markedly reduced accuracy in adversarial scenarios, while probabilistic planners such as QMDP, POMCP, and DESPOT maintain higher robustness.}
    \label{fig:heatmap}
\end{figure}

Figure~\ref{fig:heatmap} provides a complementary visualization of planner performance across all 60 generated scenarios. 
The heatmap displays action accuracy for each planner (columns) across individual scenarios (rows), with a dashed line separating the first twenty adversarial cases from the remaining non-adversarial ones. 
As expected, the FSM baseline exhibits notably lower accuracy in the adversarial region, reflecting its difficulty in handling simultaneous arrivals, non-compliant agents, and sudden pedestrian appearances. 
In contrast, probabilistic planners such as QMDP, POMCP, and DESPOT maintain higher and more stable accuracy across both regions, though some variability remains. 
This visualization highlights how adversarial scenarios effectively expose planner weaknesses, providing a clear distinction between simplistic rule-based methods and uncertainty-aware approaches.

\begin{figure}[h]
    \centering
    \includegraphics[width=0.85\linewidth]{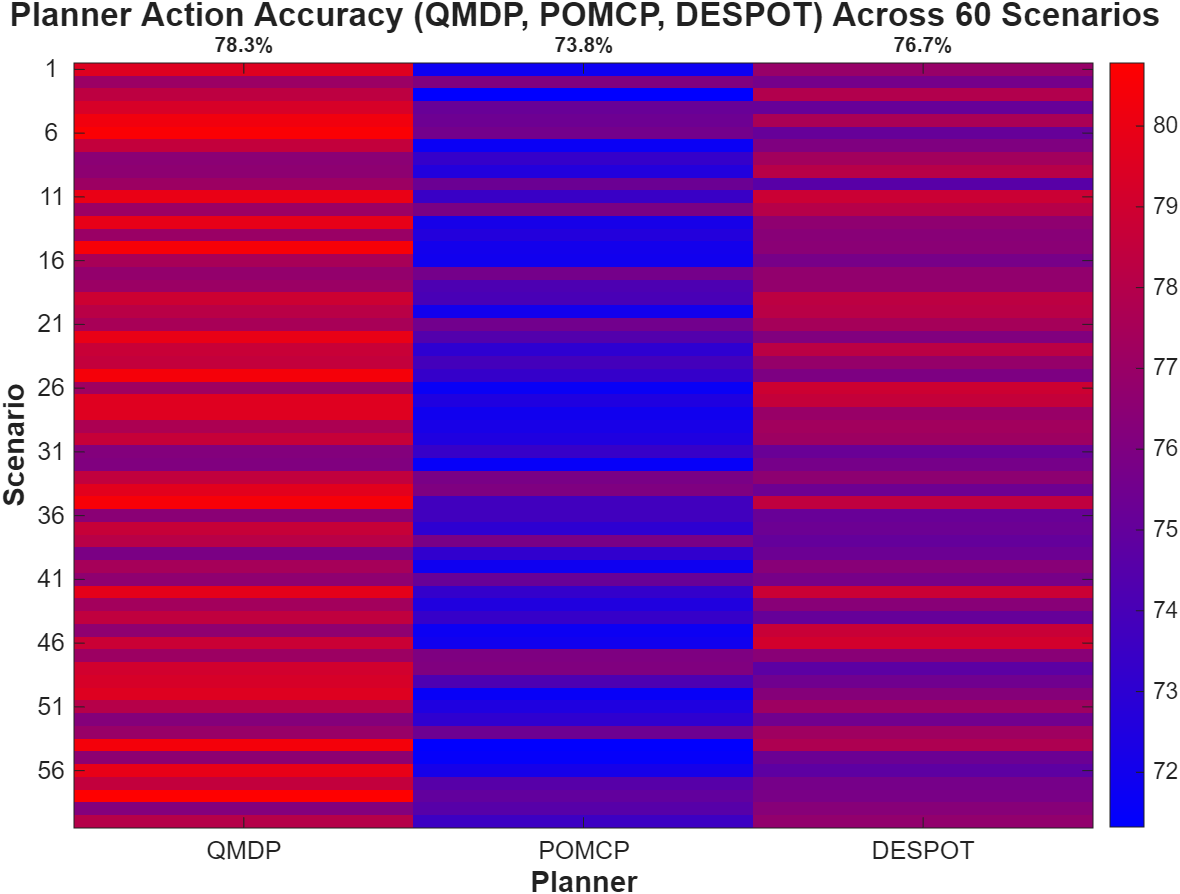}
    \caption{A view of probabilistic planner performance (QMDP, POMCP, DESPOT) across 60 scenarios. The first twenty rows correspond to adversarial scenarios. By focusing on these planners, differences in accuracy are more apparent, while FSM is excluded due to consistently lower performance in adversarial cases.}
    \label{fig:heatmap}
\end{figure}

\begin{figure}[H]
    \centering
    \includegraphics[width=0.9\linewidth]{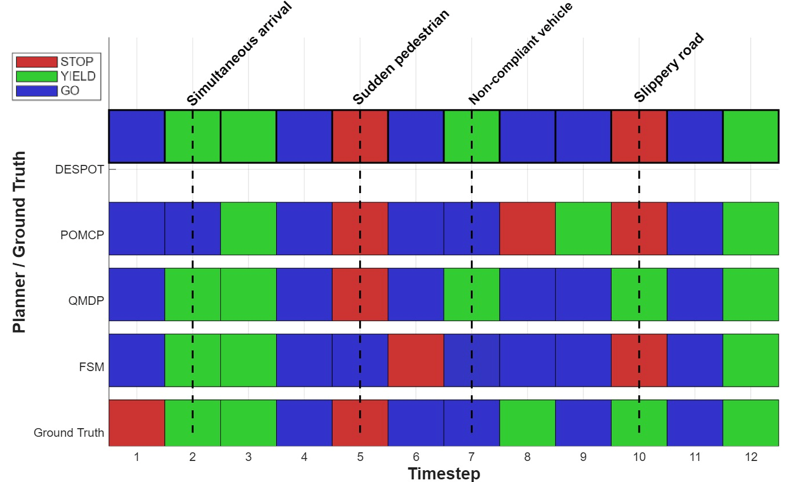}
    \caption{Planner decision trajectories for a representative adversarial scenario. The bottom row shows the ground truth actions (STOP, YIELD, GO) at each timestep, while the rows above show each planner's selected actions. Deterministic FSM frequently deviates from the ground truth, whereas uncertainty-aware planners (QMDP, POMCP, DESPOT) track the correct actions more closely. Differences among probabilistic planners highlight trade-offs between conservative and adaptive behavior under partial observability.}
    \label{fig:planner_trajectories}
\end{figure}

This visualization provides intuition behind the varying performance of planners. FSM, being deterministic, fails to adjust when multiple vehicles arrive simultaneously or when pedestrians appear unexpectedly, resulting in frequent incorrect actions. In contrast, probabilistic planners account for uncertainty and partial observability, allowing them to adapt their decisions as new information emerges. Among uncertainty-aware planners, differences arise due to how each planner models and propagates uncertainty: for example, POMCP’s particle-based sampling allows it to react flexibly to sudden changes, DESPOT balances exploration and risk through sparse tree evaluation, and QMDP applies a simpler belief-based approximation that may overlook rare but critical events. This illustrates that while uncertainty-aware frameworks outperform deterministic approaches overall, the specific strategy for handling uncertainty critically influences performance in complex, adversarial scenarios.

\end{document}